\documentclass[runningheads]{llncs}
\usepackage[T1]{fontenc}
\usepackage{array}
\usepackage{color}
\usepackage{graphicx}
\usepackage{hyperref}
\usepackage[capitalize, noabbrev]{cleveref}
\usepackage{listings}

\usepackage[backend=bibtex]{biblatex}
\usepackage{svg}
\usepackage[inline]{enumitem}
\usepackage[nolist,nohyperlinks]{acronym}
\begin{acronym}
    \acro{AI}{artificial intelligence}
    \acro{AIA}{Artificial Intelligence Act}
    \acro{AIaaS}{AI as a service}
    \acro{CNIL}{Commission Nationale de l’Informatique et des Libertés}
    \acro{DSR}{data subject right}
    \acro{EDPB}{European Data Protection Board}
    \acro{GDPR}{General Data Protection Regulation}
    \acro{LLM}{large language model}
    \acro{ML}{machine learning}
    \acro{NLP}{natural language processing}
    \acro{RAG}{retrieval-augmented generation}
    \acro{ROA}{right of access}
    \acro{RTBF}{right to be forgotten}
    \acro{RTE}{right to erasure}
    \acro{RTR}{right to rectification}
    \acro{RTROP}{right to restriction of processing}
    \acro{SISA}{Sharded, Isolated, Sliced, and Aggregated}
\end{acronym}
\usepackage{multirow}
\begin{document}
\title{Short paper: Models in the dark -- Rectification and erasure under GDPR in ML supply chains
}
\titlerunning{Rectification and erasure under GDPR in ML supply chains}

\author{
Henrik Graßhoff\orcidID{0009-0002-2310-6820} \and
Malte Hansen\orcidID{0000-0003-4622-3819} \and 
Meiko~Jensen\orcidID{0009-0003-2397-9813} \and
Sara Ramezanian\orcidID{0000-0002-5526-0817}
}
\authorrunning{Graßhoff et al.}
\institute{Karlstad University, Karlstad, Sweden \\
\email{\{henrik.grasshoff|malte.hansen|meiko.jensen|sara.ramezanian\}@kau.se}}
\maketitle

% Included for preprint of Accepted Version
\let\svthefootnote\thefootnote
\let\thefootnote\relax
\footnotetext{This version of the article has been accepted for presentation at \emph{Annual Privacy Forum 2026} and for publication, having undergone peer review, but is not the Version of Record and does not reflect post-acceptance improvements, or any corrections.}
\let\thefootnote\svthefootnote
% Delete this block for camera-ready submission

\begin{abstract}
The rights to rectification and erasure, as established under the General Data Protection Regulation (GDPR), are central to protecting individuals’ privacy. However, their effective enforcement in machine learning (ML) systems remains challenging. Existing work has largely addressed these rights from either a legal or a technical perspective in isolation and disregards the fact that models are produced in complex supply chains involving multiple actors across development, distribution, and deployment. This paper presents a holistic survey of challenges in implementing the rights to rectification and erasure in ML models. Drawing on academic literature and guidance from data protection authorities, we find that many GDPR requirements cannot yet be technically met in practice. Our findings further suggest that issues arising in ML supply chains are insufficiently addressed in research. To tackle this gap, we introduce the notion of \emph{models in the dark}~-- derived models created further downstream in an ML chain without sufficient transparency or traceability~-- and analyse the urgent challenges posed by this phenomenon. By adopting an interdisciplinary perspective, this work contributes to bridging the gap between legal requirements and the technical implementation of data subject rights in ML, ultimately supporting the development of trustworthy artificial intelligence.

\end{abstract}
\acresetall

\section{Introduction}

The \ac{GDPR}~\cite{gdpr} grants individuals a number of \acp{DSR} to exercise control over their personal data. Under Art.~16 and~17, data subjects can request that a controller rectifies inaccurate personal data or, in specific cases, deletes it. As a growing number of \ac{ML} models are trained on personal data~\cite{king_user_2025}, implementing \acp{DSR} in \ac{ML} models has increasingly attracted attention from researchers and data protection authorities~\cite{ konferenz_der_unabhangigen_datenschutzaufsichtsbehorden_des_bundes_und_der_lander_orientierungshilfe_2025}.

The implementation of the \acp{DSR} in \ac{ML} cannot rely on traditional legal and technological conceptions. Implicit binary assumptions underpinning data protection~-- such as that data is either deleted or not~-- are contested by the stochastic nature of \ac{ML} models that process data based on learnt statistics and inference. Technical procedures which are trivial for databases become complex and resource-consuming in \ac{ML} models. The popularisation of \acp{LLM} has moreover led to a ubiquity of \ac{ML}.

Research in this area, however, often misses the bigger picture. Most works focus on the \acl{RTE} and specific machine unlearning algorithms, whereas only a few~\cite{feretzakis_gdpr_2025, juliussen_algorithms_2023, zhang_right_2025, hawkins_decision-making_2024} explicitly consider \ac{GDPR} requirements. Moreover, we are unaware of any study that accounts for modern \ac{ML} models as products of complex \emph{\ac{ML} supply chains}, involving multiple actors across stages from data collection to deployment~\cite{cobbe_understanding_2023}.

To address this gap, this article investigates the following research questions:
\begin{enumerate}[label=\textbf{RQ\,\arabic*.},ref=RQ\arabic{enumi},leftmargin=*]
    \item \emph{What legal, technical, and operational challenges arise in implementing the \ac{GDPR}'s \acl{RTR} and \acl{RTE} in \ac{ML}, and how can they be technically addressed?} \label{rq: challenges}
    \item \emph{What challenges arise specifically in \ac{ML} supply chains?}
\end{enumerate}
To this end, we examine and expand on existing technical and legal academic works and guidance issued by data protection authorities. This literature review~\cite{grant_typology_2009} focuses on the rights to rectification and erasure since the challenges we observe for those are comparable, but different from e.g. the right to access or the right to restriction of processing (Art.~15, 18~\ac{GDPR}).

The contributions of this paper are:
\begin{enumerate*}[label=(\roman*)]
    \item We provide a taxonomy of challenges which impede an effective enforcement of the rights to rectification and erasure;
    \item we analyse legal requirements for the two rights in \ac{ML} contexts and assess technical solutions against these;
    \item we introduce the notion of \emph{models in the dark} and demonstrate its importance for future research and discussions.
\end{enumerate*}
By bridging technical and legal literature, we aim to contribute to establishing effective procedures and guidelines for \acp{DSR} enforcement in \ac{ML} contexts.

The rest of the paper is structured as follows. \cref{sec: DSR} outlines the rights to rectification and erasure and illustrates a typical data flow in \ac{ML} supply chains. We then present challenges arising within a model in \cref{sec: model-intrinsic} and within supply chains in \cref{sec: supply chains}. The paper concludes with \cref{sec: conclusion}.

\section{Data subject rights in machine learning}
\label{sec: DSR}

\subsection{The rights to rectification and erasure}

The \ac{GDPR}~\cite{gdpr} introduces several \acfp{DSR} to strengthen individuals' control over their personal data, including the \ac{RTR} and the \ac{RTE}. Under Art.~16, individuals may request that a controller rectify inaccurate personal data without undue delay. The \acl{RTE}, also known as the \ac{RTBF}, entitles individuals to request that a controller delete their personal data without undue delay (Art.~17~\ac{GDPR}). In this context, the controller is the organisation determining the purposes and means of processing (Art.~5~\ac{GDPR}). Where data have been shared with third parties, the controller must notify them of the request, unless this is impossible or requires disproportionate effort (Art.~19~\ac{GDPR}). If the data were made public, the controller should take reasonable steps to inform the recipients, considering available technologies and implementation costs (Art.~17(2)~\ac{GDPR}).

Scholars have questioned whether, and to what extent, \aclp{DSR} can be reasonably interpreted in the context of \ac{ML}~\cite{juliussen_algorithms_2023,villaronga_humans_2018,zhang_right_2025}. While some see its technology-neutrality (Recital~15~\ac{GDPR}) as a strength for regulating \ac{ML}~\cite{ufert_ai_2020,smuha_ai_2025}, others criticise the \ac{RTE} for relying on ``outdated metaphors like remembering and forgetting'' which ``[are] unique to human minds only and [do] not necessarily translate to the AI/machine learning era''~\cite{villaronga_humans_2018} and that ``it is necessary to interpret the right to erasure in a ML context''~\cite{juliussen_algorithms_2023}.

\subsection{Personal data in the \ac{ML} supply chain}

As a data-driven technology, personal data handling is a key concern in \ac{ML} applications. Where models contain personal data, they must comply with the \ac{GDPR}~\cite{gdpr}. As highlighted by the \ac{EDPB}, these requirements cannot easily be disregarded, since ``even when an AI model has not been intentionally designed to produce information relating to an identified or identifiable natural person from the training data, information from the training dataset, including personal data, may still remain `absorbed' in the parameters of the model''~\cite[p.~13]{european_data_protection_board_opinion_2024}. It is therefore necessary to examine where in the \ac{ML} supply chain personal data is processed and which actors are involved. 

\begin{figure}[ht]
    \centering
    \includegraphics[width=.75\linewidth]{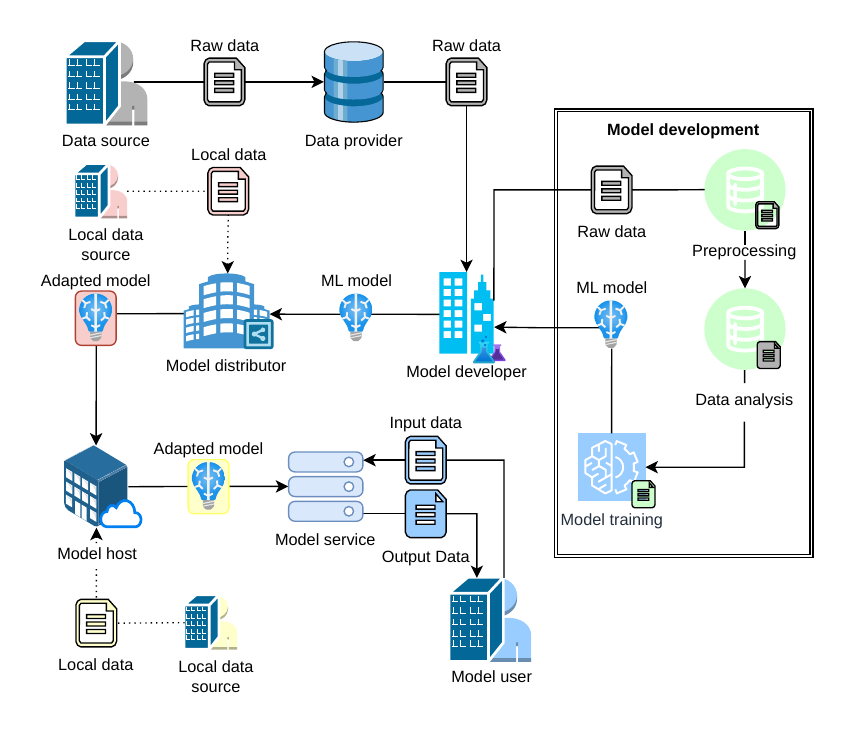}
    \caption{Data flows between different actors in a simplified \ac{ML} supply chain. Dashed lines indicate optional model adaptions.}
    \label{fig:actors}
\end{figure}

Following the \ac{ML} workflow of Glerean~\cite{glerean_fundamentals_2025}, we derive the following general data flows between actors in \ac{ML} supply chains (see \cref{fig:actors}):
\begin{enumerate}[label=(\arabic*)]
    \item Initially, raw data is collected from diverse \emph{Data sources}~\cite{gao_pile_2021}, including customer databases, patient records, or web scraping from individuals or external organisations. Data acquisition is typically performed by the organisation itself or by third parties such as vendors or aggregators, which we collectively refer to as \emph{Data provider}.
    \item \label{item: data pre-processing} The next step is data pre-processing, where raw data is prepared for model development. Data sanitisation techniques, such as removing direct and quasi-identifiers or applying differential privacy (e.g.~\cite{sinha_vaultgemma_2025}), may be used to reduce the presence of personal data. Additional steps include filtering, normalisation, and imputation, during which data values may be modified, potentially affecting the \ac{RTE}. \item During data analysis, pre-processed data is further prepared for \ac{ML} through extraction, transformation, labelling, augmentation, or synthesis. Although less common, these operations may generate new personal data by combining previously unlinked information.
    \item \label{item: training} The model is then trained using the prepared data, typically split into training, testing, and validation subsets. Based on training results, model weights are adjusted, and the resulting \ac{ML} model is deployed. In our generalised scenario, steps \ref{item: data pre-processing}--\ref{item: training} are performed by a single actor, the \emph{Model developer}. However, these tasks may also be outsourced, or, as in federated learning (see~\cite{zhang_survey_2021}), carried out collaboratively by multiple organisations.
    \item The resulting \ac{ML} model is then distributed. The \emph{Model developer} delivers it to the \emph{Model distributor}, which may be the same organisation, or one for which the model was developed. The \emph{Model distributor} may either pass the model to an external \emph{Model host} or act as the host itself. Both actors may further fine-tune the model using local \emph{Data sources}.
    \item Finally, the \emph{Model host} offers the model to a \emph{Model user} as an \ac{ML} service. During usage, the user submits input data, which is processed to produce output. The \emph{Model user} may be any entity, from an individual using a chatbot to an organisation classifying customers. Consequently, the input data may contain new personal data and, depending on the architecture, may be stored locally within the \emph{Model user}'s instance of the \ac{ML} service or transferred to the \emph{Model host}. Output data may also include personal data inferred from the model and potentially traceable to original training data. In some cases, both input and output data are reused for further training.
\end{enumerate}

\subsection{Scope of this paper}

An \ac{ML} supply chain thus leads to a multitude of data objects held by different controllers. The \ac{RTR} and \ac{RTE} can just as well be filed against data used for model training or adaptation (see~\cref{fig:actors}) as it can be against the models themselves~\cite{cnil_ensuring_2026, european_data_protection_board_opinion_2024, konferenz_der_unabhangigen_datenschutzaufsichtsbehorden_des_bundes_und_der_lander_orientierungshilfe_2025}. While the precise legal roles (processor, controller, or joint controllers) depend on the specific use case~\cite{cnil_determining_2024}, we assume for the remainder that a data subject files a valid \ac{RTR} or \ac{RTE} request \emph{against a data controller's model} at an arbitrary stage in the supply chain.
In the next sections, we investigate two types of challenges which occur:
\begin{enumerate*}[label=(\arabic*)]
    \item model-intrinsic challenges, which stem from the training and black-box nature of the models themselves, and
    \item supply chain challenges, that arise from the magnitude of actors involved.
\end{enumerate*}
We acknowledge that human factors play a crucial role for empowering individuals in exercising their data subject rights~\cite{bowyer_human-gdpr_2022}, but consider human aspects out of scope for this paper and worthy of a dedicated work on its own.

\section{Model-intrinsic challenges}
\label{sec: model-intrinsic}

\ac{ML} models may memorise personal data contained in their training data and have been shown to leak these verbatim or approximately~\cite{carlini_extracting_2021, carlini_extracting_2023, huang_are_2022, ippolito_preventing_2022, rigaki_survey_2024}. Yet, this information is not stored explicitly within the model; rather, it is implicitly encoded in the model's parameters, which gives rise to several unique issues.

\subsection{Necessity to assess models, not training data}

A key insight is that a model is not equivalent to its training data. It may fail to memorise individual data points due to limited parameter capacity or stochasticity in training, where data are randomly sampled, ordered, and processed~\cite{bourtoule_machine_2021}. Conversely, models may also generate information not present in the training data, either through hallucination of incorrect or fabricated outputs~\cite{ji_survey_2023, kelsey-sugg_ai_2024} or by inferring new information from training data.

Any assessment of what a model has stored must therefore be based on the model itself rather than its training data~\cite{cnil_ensuring_2026}. This argumentation also applies to rectification and erasure methods: a technique effective for one model will not necessarily be applicable to other models trained on the same~-- let alone a similar~-- dataset.

\subsection{Identification of the data subject within a model}
\label{sec: identification}

Quantifying the influence of a specific training data point on a model remains an open problem~\cite{bourtoule_machine_2021, cnil_ensuring_2026, kesa_artificial_2020}, making unambiguous and complete identification of data subjects difficult, if not impossible.

When a controller receives a \ac{DSR} request, it is obliged to identify the data subject within the model, which involves to determine what personal data the model has stored about the latter~\cite{cnil_ensuring_2026, european_data_protection_board_guidelines_2023}. As stressed above and supported by the \ac{CNIL}~\cite{cnil_ensuring_2026}, the controller cannot base this assessment solely on the training data, but has to verify the stored data within the model itself on a case-by-case basis.

For generative models such as \acp{LLM}, the most mature technical identification approach today is \emph{privacy audits}. These often involve inserting ``canaries''~\cite{carlini_secret_2019} into the training dataset: random information pairs later tested for leakage through methods such as \emph{membership inference attack}~\cite{hu_membership_2022}. However, some studies indicate that many methods struggle to identify memorised data reliably and remain sensitive to canary and prompt design~\cite{ippolito_preventing_2022, lu_fantastically_2022}. Their reliance on control over training data and processes limits applicability to the model developer, who may not be the addressed controller (see~\cref{sec: addressing}). Lastly, quantifying a model's overall tendency for privacy leakage may not meet the needs of addressing an individual data subject's case.

More generally, it is debatable to what extent a controller can be obliged to audit its model in light of Art.~11 \ac{GDPR}. This provision prohibits controllers from maintaining, acquiring, or processing additional information solely to identify a data subject. Privacy audits may constitute such processing. Moreover, any model attack inherently risks infringing individual privacy. Where controllers do not hold the training data, requiring black-box auditing may conflict with the rationale of Art.~11: enabling one data subject’s rights should not compromise the privacy of others.

Facing these difficulties, the controller could be inclined to argue that it cannot identify the data subject and refuse to comply with the \ac{DSR} request with reference to Art.~12(2)~\ac{GDPR}. The wording of the \ac{CNIL} guidelines in fact suggests that ``it does not seem possible today, in the general case, for a controller to identify … a particular person'' and the controller could then ``demonstrate that it is unable to identify persons within its model'', except for rare cases in which the data is explicitly encoded in the model weights~\cite{cnil_ensuring_2026}. Even though this argumentation may appear reasonable given the current legislation and available technologies, it could function as a dangerous loophole for data controllers to escape \ac{DSR} requests if interpreted too broadly.

\subsection{Technical data rectification and erasure in a model}
\label{sec: unlearning}

Fully \emph{retraining} a model from scratch on revised training data is widely considered to be the most \ac{GDPR}-compliant way to implement the \ac{RTR} and \ac{RTE}~\cite{cnil_ensuring_2026, feretzakis_gdpr_2025}, but very resource-intensive~\cite{cao_towards_2015}. In this regard, conflicting opinions exist on whether a data subject can, in fact, request a full retraining from the data controller. Juliussen et al.~\cite{juliussen_algorithms_2023} argue that as ``the only current machine unlearning method with a theoretically proven guarantee is … full model retraining, … the request for full model retraining has a legal basis in Art.~17 of the \ac{GDPR} and should be implemented for full \ac{GDPR} compliance.'' In contrast, high implementation costs may be a reason for the controller to refuse the \ac{DSR} request according to \ac{CNIL}: ``The retraining of the models comes at a very high cost. The request for erasure may in principle be rejected''~\cite{cnil_ensuring_2026}.

Less resource-intensive approaches aim to \emph{unlearn} a specific subset of the training data from a learnt model, offering a trade-off between erasure strictness, model performance, and resource consumption~\cite{liu_survey_2025, nguyen_survey_2025, shaik_exploring_2025}. These methods are first and foremost intended for erasing data from a model, but can also be used for rectification through a combination of erasure and relearning. While being a very active field of research and arguably the best known feasible approach to implement the \ac{RTR} and \ac{RTE} today, they introduce a number of issues.

Unlearnt models typically perform worse than retrained ones, with performance potentially deteriorating exponentially as more data is removed~\cite{nguyen_variational_2020, nguyen_survey_2025}. This phenomenon, known as \emph{catastrophic unlearning}, limits the applicability of machine unlearning. A controller may have little incentive to unlearn data if it renders the model ineffective. Since the controller's freedom to conduct a business must be respected (Recital~4~\ac{GDPR} and \cite{cnil_ensuring_2026}), it is as of today unclear to what extent catastrophic unlearning can justify non-compliance with a \ac{DSR} request. Moreover, controllers of high-risk \ac{AI} systems under the \ac{AIA}~\cite{aia} may be further reluctant as such systems must maintain appropriate levels of accuracy (Art.~15(1) \ac{AIA}). Finally, since unlearning may introduce unfairness by disproportionately affecting subgroups~\cite{nguyen_survey_2025, zhang_be_2024}, it may conflict with the accuracy and fairness principles in Art.~5~\ac{GDPR}.

Another important aspect is the verifiability of rectification or erasure~. \emph{Exact unlearning} algorithms provide formal guarantees that the model behaves as if it had never encountered the data~\cite{chen_survey_2025}. \emph{Approximate unlearning} algorithms suppress the influence of a subset of the training data without removing it from the model~\cite{chen_survey_2025}. While significantly less resource-intensive~\cite{xu_machine_2024}, approximate methods may leave residual effects~\cite{nguyen_survey_2025}. Controllers must then verify the unlearning, yet no standard mechanisms exist, and proposed approaches remain limited~\cite{xue_towards_2026}.

Ironically, machine unlearning itself can also expose individuals to different privacy harms, exact and approximate algorithms alike~\cite{chen_survey_2025, liu_survey_2025}. The fundamental idea is that an adversary who can query a model before and after the unlearning can leverage this additional information to their advantage. Past studies have shown higher success rates for membership inference attacks~\cite{chen_when_2021} or, more severely, \emph{data reconstruction attacks}, in which the unlearnt training data could be recovered from the models with high confidence~\cite{bertran_reconstruction_2024}.

\subsection{Inference-time interventions}

Alternatively to changing the trained model parameters, \emph{in\-fe\-r\-ence-time interventions}~\cite{li_inference-time_2023} modify the input/output of a generative model during interactions. Such mechanisms are widely deployed for detecting and suppressing prompts or responses that are considered harmful~\cite{dong_safeguarding_2025, fu_sanitize_2025, gandikota_erasing_2023, teodora_musatoiu_how_2024} and have also been proposed as alternatives to unlearning~\cite{pawelczyk_-context_2023, thaker_guardrail_2024}. Yet, it is acknowledged that the measures ``would not satisfy definitions of unlearning''~\cite{thaker_guardrail_2024} since the data is not technically rectified or erased from the model. Given the documented circumventions of inference-time interventions~\cite{hackett_bypassing_2025,}, it is not obvious how these could play a vital role in the \ac{RTR} and \ac{RTE} implementation apart from emergency measures when all other measures prove disproportionate~\cite{cnil_ensuring_2026}.

\subsection{Continually learning models}

Models may learn continually from a stream of training data, gradually accommodating new knowledge over time while maintaining their already acquired abilities~\cite{wickramasinghe_continual_2024}. This learning approach (also known as online, incremental, or lifelong learning~\cite{wang_comprehensive_2024, wickramasinghe_continual_2024}) does not produce a ``final model'', but a ramified directed graph of continuously updated models, each of which depends on its predecessors and therefore, implicitly, on all previous training data~\cite{bourtoule_machine_2021}. A \ac{DSR} request filed for one model in this graph would inevitably affect all subsequent models, potentially causing a cascade of necessary rectifications or erasures, each of which is subject to the challenges described in the previous subsections.

\section{Supply-chain challenges}
\label{sec: supply chains}

\subsection{The notion of \emph{models in the dark}}

\begin{figure}[t]
    \centering
    \includegraphics[width=1\linewidth]{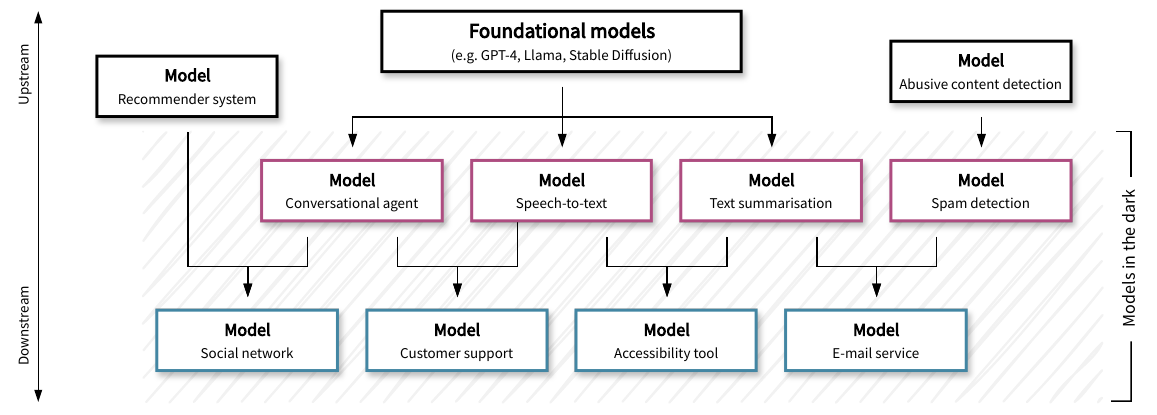}
    \caption{Models in the dark in simple \ac{ML} supply chains (adapted from~\cite{hopkins_ai_2024})}
    \label{fig:mitd}
\end{figure}

\ac{ML} models are rarely designed and used in a vacuum. They are the result of a larger supply chain which involves multiple actors in all steps from problem formulation to data collection and curation, model training, evaluation, distribution, deployment, usage, and monitoring~\cite{brown_allocating_2023, cooper_accountability_2022, hopkins_ai_2024, hopkins_recourse_2025}. During this process, models are constantly adapted, combined, and redistributed~\cite{cobbe_artificial_2021, noauthor_hugging_nodate}, creating a complex network of subsequent downstream derivatives of the original model as illustrated in \cref{fig:mitd}. Whenever an upstream model contains personal data, this information is also propagated and may ultimately, through cascading effects, appear in an unbounded number of downstream models.

Still, actors typically lack knowledge of other participants more than one ``hop'' away in the supply chain~\cite{cobbe_understanding_2023}. In particular, a data subject interacting with a downstream model as an end user is usually unaware of upstream predecessors. While they may perceive the \ac{AI} as a stand-alone product, it is in fact the result of multiple contributing models. We therefore refer to these as \emph{models in the dark}.\footnote{Inspired by the Persian expression \emph{army in the dark}, referring to unseen or uncredited contributors.} Each such model, i.e. every derivative model built from an upstream one, creates a further instance of personal data at an additional actor.

The clear distinction between upstream and downstream and the mental model of linear supply chains are too simple to capture the complexity and dynamics of reality. Modern \ac{ML} supply chains are characterised by fluid interrelations of vertically and horizontally integrated actors~\cite{attard-frost_ethics_2025, cobbe_understanding_2023, hopkins_recourse_2025} where the data and model flows between actors are much more complex than~\cref{fig:mitd} suggests, flows can even be circular, and trust relationships are diverse~\cite{balayn_unpacking_2025}. The notion of models in the dark is agnostic to the exact actor constellation as it refers to \emph{all} derivatives of a model. For the sake of simplicity, we adhere to the picture of linear downstream flows for the rest of the section.

\subsection{Locating controllers under supply chain opacity}
\label{sec: addressing}

Before submitting a \ac{DSR} request, a data subject does not need to determine whether a model contains their personal data. However, in practice, they may lack information about whether and how their data is present in an \ac{ML} model, complicating the exercise of their rights. To find out whether a model contains their data, data subjects typically rely on their interaction with the model~\cite{cnil_ensuring_2026}. Black-box methods without access to model specifications remain limited and may lead to incorrect conclusions that a model does not store their data.

In addition, lacking knowledge about other controllers in the supply chain, a data subject may submit a request at an arbitrary point. If the controller is far upstream, personal data may have been propagated to an unknown number of downstream models in the dark (see~\cref{fig:mitd}). Far downstream, the model may instead have inherited data from upstream predecessors. In both cases, even if the addressed controller rectifies or erases the data, other models may retain it, limiting effectiveness. It may therefore be necessary in many cases to ensure rectification or erasure across all relevant models in the dark.

\subsection{Updating models along the supply chain}

One potential measure to address models in the dark is to propagate an \emph{updated} version of the model with changed model weights through the supply chain.
Depending on whether, how, and to whom the controller makes the model available, different legal obligations apply:
\begin{enumerate*}[label=(\arabic*)]
    \item \label{item: no distribution} If the controller does not distribute the model, but offers it as a service~\cite{cobbe_artificial_2021}, it needs to be updated only at the controller's site.
    \item \label{item: distribution no public} If the controller provides the model itself to third parties without making it publicly available, they are obliged to undertake proportionate efforts to inform all recipients of the \ac{RTR} or \ac{RTE} request (Art.~19~\ac{GDPR}) if possible. According to the assessment of \ac{CNIL}, the controller would then have to forward an updated version downstream and, if necessary, contractually restrict the recipient's use to the new version~\cite{cnil_ensuring_2026}.
    \item \label{item: model public} If the controller makes the model publicly available, Art.~17(2)~\ac{GDPR} applies for the \ac{RTE} and mandates that the controller should take reasonable actions to inform controllers of the erasure request, considering available technology and the cost of implementation~\cite{juliussen_algorithms_2023}. A comparable obligation does not exist for the right to rectification.
\end{enumerate*}

From a practical perspective, case~\ref{item: no distribution} requires a model update at a single stage of the supply chain and leads to the challenges discussed in \cref{sec: model-intrinsic}. Case~\ref{item: distribution no public} imposes a potentially acceptable burden to inform selected recipients, but distributing updates to all may require substantial network traffic given the multi-gigabyte size of modern models~\cite{deepseek-ai_deepseek-v3_2025}. We are not aware of any established techniques to systematically reduce bandwidth for model updates. In case~\ref{item: model public}, it is nearly impossible to inform all parties who have downloaded the model (for instance, Llama has been downloaded over one billion times~\cite{meta_celebrating_2025}). As a remedy, published models could include an update mechanism enabling distribution akin to software updates. However, bandwidth constraints, limited enforcement, and the rise of open-weight models~\cite{noauthor_hugging_nodate} remain obstacles.

All of these considerations build upon the premise that the controller takes care of the update (distribution) itself. In some cases, for example, when the controller is not in possession of sufficient resources, rectification or erasure might be more effectively carried out further downstream. Standardised coordination procedures could, in principle, allow to establish favourable places in the \ac{ML} supply chain to execute the model updating.

Lastly, updating downstream models might not always be effective or desirable. When the number of models in the dark grows, the link between a model and its derivatives becomes less obvious as more and more models contribute to the models further downstream. The sole propagation of a model update does therefore not assure the rectification and erasure in all derivations. Referring back to \cref{sec: unlearning}, updating a model can also infringe on the data subjects' privacy, rendering large-scale model updates potentially undesired.

\section{Conclusion and future work}
\label{sec: conclusion}

In this paper, we surveyed academic literature and data protection authority guidelines to identify challenges faced by the \acl{RTR} and \acl{RTE} in \ac{ML} systems. The result is a taxonomy of two domains: model-intrinsic and supply-chain challenges. For each domain, we identified key challenges, corresponding technical approaches, and limitations, extending existing literature.

For model-intrinsic challenges, key issues include limited methods for reliably identifying personal data within models and the shortcomings of current unlearning techniques. Supply-chain challenges remain underexplored: complex actor constellations create \emph{models in the dark}, complicating both the identification of responsible controller(s) and the distribution of resulting model updates.

Overall, both the technological implementations for the \ac{RTR} and \ac{RTE} and the supply chains surrounding them contain challenges that make the effective enforcement of either infeasible. Hence, the proper enforcement of \ac{RTR} and \ac{RTE} in \ac{ML} contexts raises a lot of open issues that require further research in several areas to assert that data subjects keep the ability to enforce their rights in the light of the rise of \ac{ML}.  

In future work, we plan to develop and evaluate strategies to address the challenges introduced in this paper. Specifically, we plan to conduct technical experiments to evaluate the effectiveness of unlearning and related techniques for the \ac{RTR} and \ac{RTE} and to minimise network traffic during model distribution. To elucidate the uncertainty surrounding models in the dark, we will investigate and develop communication flows and data trails for \ac{GDPR}-compliant and traceable data sharing and model distribution in \ac{ML} supply chains. Another area of challenges, beyond this paper’s scope, concerns the training process. As highlighted in the data flow in \cref{sec: DSR}, data is transformed multiple times during model development. The implications for the \ac{RTR} and \ac{RTE} require further attention, particularly in distributed training settings such as federated learning.

\subsubsection*{AI usage disclaimer.}

The authors utilised generative AI tools for improving grammar and spelling.

\subsubsection*{Acknowledgements.}

The contribution of Sara Ramezanian and Meiko Jensen was co-funded by the Swedish Knowledge Foundation (KKS). The contribution of Malte Hansen and Meiko Jensen has been conducted (in part) within the project TRUMAN. The research leading to these results has received funding from the European Union's Horizon 2020 Research and Innovation Programme, under Grant Agreement no. 101214000.

\subsubsection*{Author contributions (\href{https://credit.niso.org/}{CRediT}).}

\textbf{Hen\-rik Graß\-hoff:} Conceptualization, Investigation, Project administration, Visualization, Writing -- original draft, Writing -- review~\& editing;
\textbf{Mal\-te Han\-sen:} Conceptualization, Investigation, Visualization, Writing -- original draft, Writing -- review~\& editing;
\textbf{Mei\-ko Jen\-sen:} Conceptualization, Project administration, Writing -- review~\& editing;
\textbf{Sa\-ra Ra\-me\-za\-ni\-an:} Conceptualization, Writing -- review~\& editing.

\printbibliography

% Included for preprint of Accepted Version
\vfill
\subsubsection*{Open Access.}
This manuscript is licensed under the terms of the Creative Commons Attribution-NonCommercial-NoDerivatives 4.0 International License (\href{https://creativecommons.org/licenses/by-nc-nd/4.0}{https://creativecommons.org/licenses/by-nc-nd/4.0}), which permits any noncommercial use, sharing, distribution and reproduction in any medium or format, as long as you give appropriate credit to the original author(s) and the source, provide a link to the Creative Commons license and indicate if you modified the licensed material. You do not have permission under this license to share adapted material derived from this manuscript or parts of it.

~\\\noindent\includegraphics[width=0.2\linewidth]{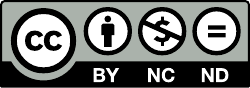}
% Delete this block for camera-ready submission

\end{document}